\documentclass[10pt,twocolumn,letterpaper]{article}

\usepackage{cvpr}
\usepackage{times}
\usepackage{epsfig}
\usepackage{graphicx}
\usepackage{amsmath}
\usepackage{amssymb}

\usepackage{booktabs}
\usepackage[breaklinks=true,bookmarks=false]{hyperref}
\makeatletter
\def\blfootnote{\xdef\@thefnmark{}\@footnotetext}
\makeatother

\cvprfinalcopy % *** Uncomment this line for the final submission

\def\cvprPaperID{76} % *** Enter the CVPR Paper ID here

\ifcvprfinal\fi
\begin{document}

%%%%%%%%% TITLE
\title{Learning Raw Image Denoising with Bayer Pattern Unification and\\Bayer Preserving Augmentation}

\author{
Jiaming Liu\textsuperscript{1}
 \quad Chi-Hao Wu\textsuperscript{1}
 \quad Yuzhi Wang\textsuperscript{2}
 \quad Qin Xu\textsuperscript{1}
 \quad Yuqian Zhou\textsuperscript{1}
 \quad Haibin Huang\textsuperscript{1}\\
 \quad Chuan Wang\textsuperscript{1}
 \quad Shaofan Cai\textsuperscript{1}
 \quad Yifan Ding\textsuperscript{3,1}
 \quad Haoqiang Fan\textsuperscript{1}
 \quad Jue Wang\textsuperscript{1}
 \\
\textsuperscript{1}Megvii Technology
\quad \textsuperscript{2}Tsinghua University
\quad \textsuperscript{3}University of Central Florida
\\
{\tt\small \textsuperscript{1}\{liujiaming,wuqihao,xuqin,zhouyuqian,huanghaibin,wangchuan,caishaofan,fhq,wangjue\}@megvii.com}\\ {\tt\small \textsuperscript{2}yuzhiwang@tsinghua.edu.cn \quad \textsuperscript{3}yf.ding@knights.ucf.edu}
}

\maketitle

\blfootnote{This work is supported by The National Key Research and Development Program of China (2018YFC0831700).}

%%%%%%%%% ABSTRACT
\begin{abstract}
In this paper, we present new data pre-processing and augmentation techniques for DNN-based raw image denoising. Compared with traditional RGB image denoising, performing this task on direct camera sensor readings presents new challenges such as how to effectively handle various Bayer patterns from different data sources, and subsequently how to perform valid data augmentation with raw images.
To address the first problem, we propose a Bayer pattern unification (BayerUnify) method to unify different Bayer patterns. This allows us to fully utilize a heterogeneous dataset to train a single denoising model instead of training one model for each pattern.
Furthermore, while it is essential to augment the dataset to improve model generalization and performance, we discovered that it is error-prone to modify raw images by adapting augmentation methods designed for RGB images.
Towards this end, we present a Bayer preserving augmentation (BayerAug) method as an effective approach for raw image augmentation.
Combining these data processing technqiues with a modified U-Net, our method achieves a PSNR of 52.11 and a SSIM of 0.9969 in NTIRE 2019 Real Image Denoising Challenge, demonstrating the state-of-the-art performance.
Our code is available at \url{https://github.com/Jiaming-Liu/BayerUnifyAug}.
\end{abstract}

%%%%%%%%% BODY TEXT
\section{Introduction}

Image denoising is one of the fundamental problems in image processing and computer vision, and restoring high quality images from extremely noisy ones remains to be challenging.
This can be even worse when it comes to images taken from mobile devices. Due to the use of relatively low-cost sensors and lenses,   images captured by mobile cameras can be severely corrupted by high level noise, especially in low-light scenarios. Many denoising methods have been proposed to address this problem, including traditional methods such as NLM~\cite{buades2005non} and BM3D~\cite{dabov2007image} as well as more recent deep neural network~(DNN) based denoising models~\cite{tai2017memnet,chen2017trainable,zhou2019awgn,jain2009natural,xie2012image,mao2016image,ulyanov2018deep}, but their performances are still far from satisfactory on mobile devices. 

\begin{figure}[t!]
\begin{center}
\includegraphics[width=\linewidth]{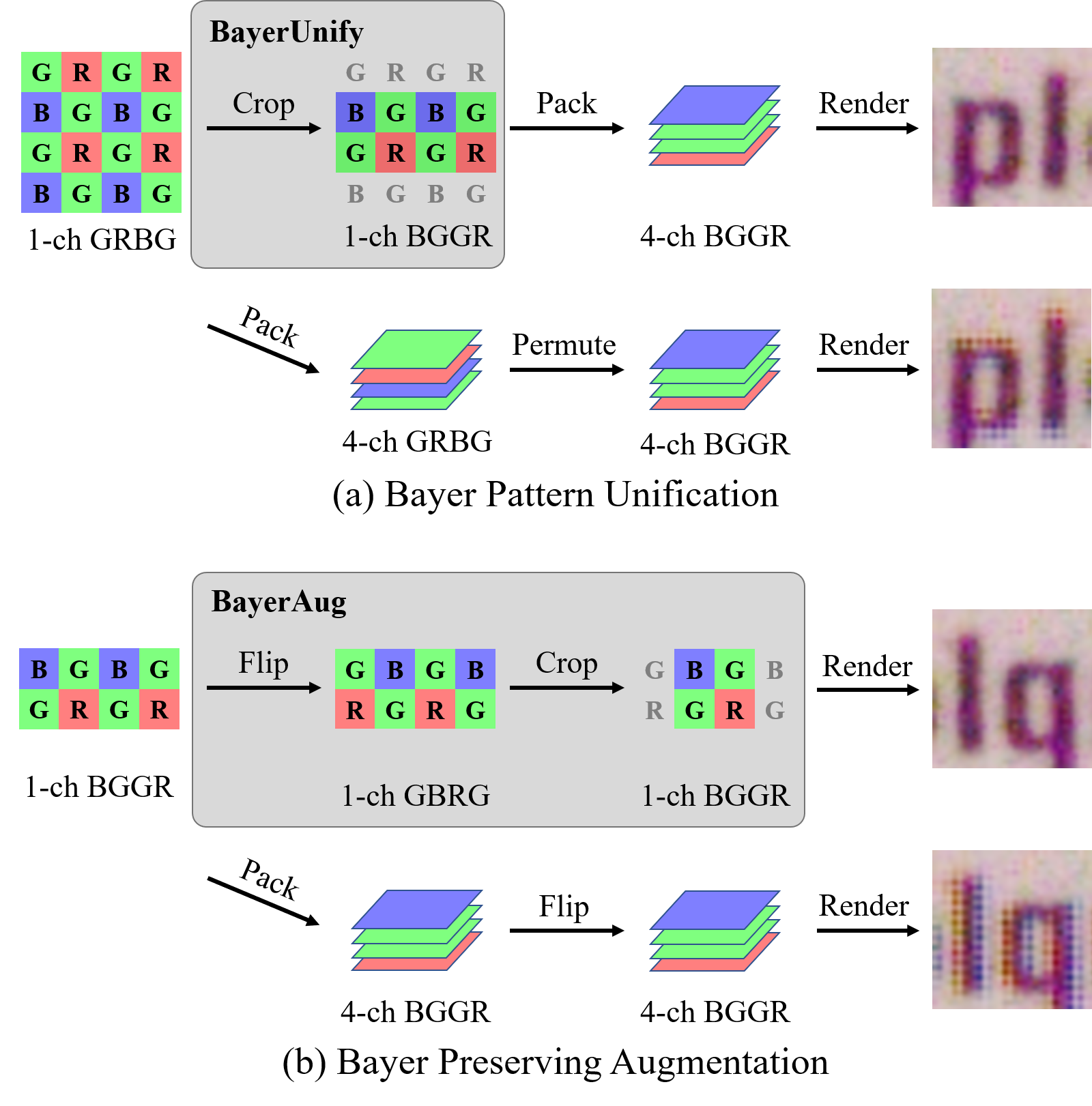}
\end{center}
   \caption{Demonstration of our proposed (a) Bayer pattern unification and (b) Bayer preserving augmentation. Our method unifies and augments the Bayer raw images without affecting the content, while improper pre-processing or augmentation would disturb the spatial relationship of the raw images and therefore result in artifacts.}
\label{fig:demo}
\end{figure}

Recently, thanks to the public raw image denoising datasets~\cite{SIDD_2018_CVPR,chen2018learning,anaya2018renoir}, denoising raw image data has received more and more attentions and has shown promising results  ~\cite{chen2018learning,hirakawa2006joint,gharbi2016deep}. Raw images are direct readings from images sensors, with camera filter arrays~(CFAs) arranged in specific patterns such as the Bayer pattern~\cite{Bayer1976color}. These digital signals are further post-processed to obtain RGB images through a complex pipeline including lens shading correction, white balancing, demosaicking, gamma correction, \etc ~\cite{hasinoff2016burst}. Therefore, original noise properties that exist in raw images are often distorted in RGB images, making the noise harder to remove afterwards. This means that there are potentially better denoising methods that can be developed on the raw image data~\cite{chen2018learning}, compared with many works done in RGB domain. In this work, we study the problem of raw image denoising using DNN, and our focus is on how to train an effective raw image denoising model by proper data pre-processing and data augmentation.  

First of all, to perform raw image denoising with DNN models, it is a common practice to pack a Bayer raw image into a 4-channel RGGB image, and feed it into neural networks~\cite{chen2018learning}. With data collected from cameras with different Bayer patterns, a simple solution is to train one model for each pattern. However, this decreases the size of the effective training set and thereby hurts the performance. To fully utilize all training data to achieve better performance, we propose a Bayer pattern unification (BayerUnify) technique to eliminate the differences among Bayer patterns. As illustrated in Fig.~\ref{fig:demo}~(a), flipping and cropping operations are employed to turn a specific CFA pattern into another one, with which we can unify all training images into the same pattern. As a result, all the training data can be used together to optimize a single model to achieve the best possible result.

Data augmentation is a common approach in deep learning to improve model performance by increasing the diversity of a training dataset. However, data augmentation of raw images is not as straightforward as that of RGB images.
An example in shown in Fig.~\ref{fig:demo}~(b). Simply flipping the packed 4-channel raw images is erroneous because it results in an image that is impossible in real world.
This phenomenon can also be found in other types of augmentation operations such as cropping, transposition, \etc. To tackle this problem, we introduce a Bayer preserving augmentation (BayerAug) technique that allows proper augmentation for raw images. As shown in Fig.~\ref{fig:demo}~(b), extra operations are required to correctly flip a raw image.

Both BayerUnify and BayerAug techniques are simple, yet effective ways for increasing the training data size and diversity for raw image denoising. We apply these techniques to train models based on our modified U-Net~\cite{ronneberger2015u}, and achieve state-of-the-art results in NTIRE 2019 Real Image Denoising Challenge (Track 1)~\cite{Abdelhamed_2019_CVPR_Workshops}. We conduct detailed experiments and analysis in Sec.~\ref{sec:experiment} to validate the proposed techniques. 

To summarize, our contributions include:

\begin{itemize}

\item  We propose a novel Bayer pattern unification technique (BayerUnify) to transform different Bayer patterns into an unified one, enabling training and inference of a single denoising model using data collected from different digital cameras.
\item  We introduce a Bayer preserving augmentation technique (BayerAug) to allow effective augmentation for raw images for further performance improvement.
\item  We build DNN models based on U-Net for raw image denoising, and apply the above techniques to achieve the best results in NTIRE 2019 Real Image Denoising Challenge (Track 1). Extensive experiments are conducted to demonstrate the effectiveness of proposed method.
\end{itemize}
%-------------------------------------------------------------------------

\section{Related Works}

Image desnoising is a fundamental low-level image processing problem that has been studied for decades. Classical approaches include non-local means (NLM) \cite{buades2005non}, sparse coding \cite{elad2006image,mairal2009non,aharon2006k}, 3D transform-domain filtering (BM3D) \cite{dabov2007image}, and others \cite{gu2014weighted,portilla2003image}. Image priors are often required to design these methods. In recent years, convolutional neural networks (CNN) allows end-to-end training of discriminative models that have achieved great success in this field. As an earlier work, \cite{burger2012image} applied multi-layer perceptron (MLP) to achieve comparable results with BM3D. Later on, more advanced network architectures are proposed. For example, \cite{zhang2017beyond} uses residual learning and batch normalization to achieve great performance improvement. \cite{tai2017memnet} applies memory blocks to tackle the long-term dependency problems in previous CNN architectures. A large number of CNN works are continuously proposed \cite{chen2017trainable,jain2009natural,xie2012image,mao2016image,ulyanov2018deep,zhang2018ffdnet,lehtinen2018noise2noise} to tackle this long-standing problem.

While there are many single image denoising approaches focusing on RGB images, raw image denoising has attracted much less attention due to the lack of training data. Some works use CNN to process the noisy raw images to produce clean RGB images \cite{chen2018learning,hirakawa2006joint,gharbi2016deep} and achieve impressive visual results. In these works, the CNN models not only deal with the denoising problem, but concurrently handle other problems such as demosaicking and so on. For pure raw to raw image denoising, it is feasible to directly apply RGB denoising methods on raw images as demonstrated in \cite{SIDD_2018_CVPR}, but the performance is limited. Recently, with more publicly available raw image datasets \cite{SIDD_2018_CVPR,chen2018learning,anaya2018renoir}, it is expected to see more efforts put on solving the raw image denoising problem. In this paper, we mainly resolve the specific issues of data pre-processing and augmentation with Bayer raw images.

\begin{figure*}[t]
\begin{center}
\includegraphics[width=0.95\linewidth]{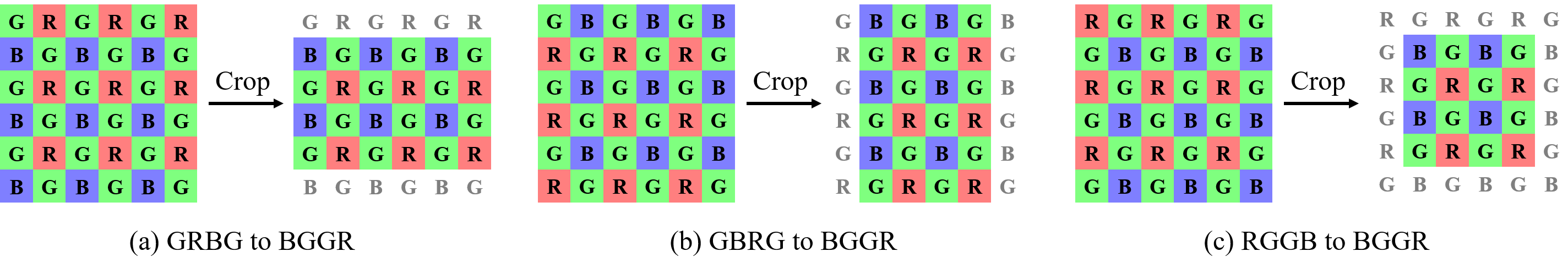}
\end{center}
   \caption{Unify Bayer pattern via cropping in the training phase.}
\label{fig:crop}
\end{figure*}

\begin{figure}[ht]
\begin{center}
\includegraphics[width=\linewidth]{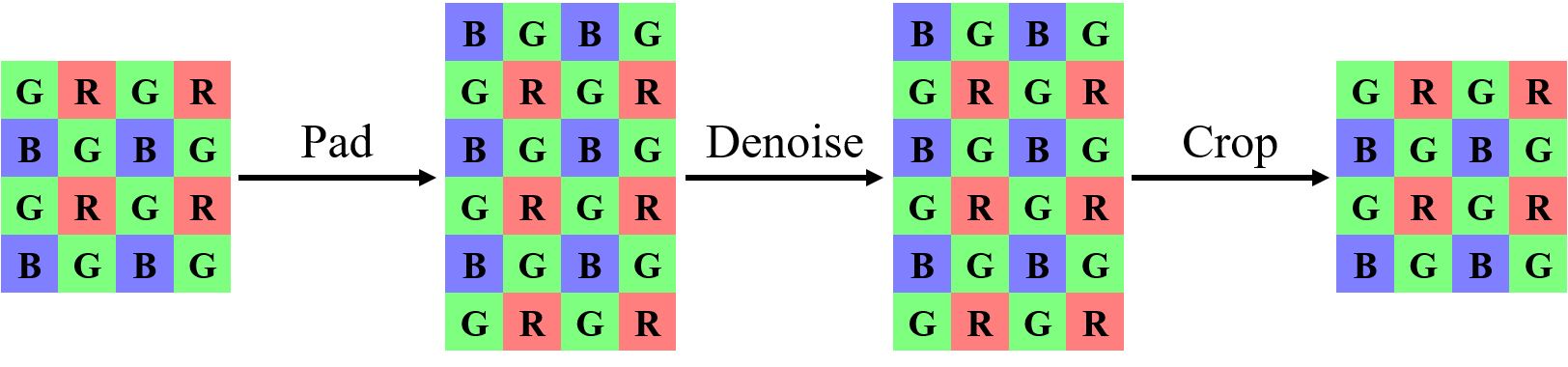}
\end{center}
   \caption{Unify Bayer pattern via padding and disunify via cropping in the testing phase.}
\label{fig:pad}
\end{figure}

\section{Proposed Method}

\subsection{Bayer Pattern Unification (BayerUnify)}
\label{sec:Bayernorm}
The Bayer patterns of raw images fall into different categories. To apply a single CNN to denoise raw images with different Bayer patterns, it is essential to align the order of the channels since different channels capture different regions of wavelength. In the meantime, the structural information laid in adjacent pixels from different channels has to be maintained. Based on these principles, we propose multiple ways to convert a raw image from one Bayer pattern to another, which are applicable to different scenarios.

\subsubsection{Training: Unify via Cropping}

In the training stage, we unify raw images with different Bayer patterns via cropping. By scarifying a minor number of pixels, it enables us to use raw images from different cameras to train a single denoising model, and thereby increases the number of available training samples.

We first introduce our notations of Bayer patterns. We represent each pattern by the sequence of its channels within each $2\times2$ block, in the order of top-left, top-right, bottom-left, and bottom-right. Typically, there are 4 possible formats, namely RGGB, BGGR, GRBG, and GBRG. For clarity, we use BGGR as the target format to illustrate our method.
 
Cropping odd number of rows or columns creates offsets which alter the Bayer pattern. As shown in Fig.~\ref{fig:crop}, cropping the first row and the last row changes a $C_1C_2C_3C_4$ image into a $C_3C_4C_1C_2$ image (e.g. GRBG to BGGR). Likewise, cropping the first and the last column alters $C_1C_2C_3C_4$ into $C_2C_1C_4C_3$ (e.g. GBRG to BGGR). These two operations together convert $C_1C_2C_3C_4$ into $C_4C_3C_2C_1$ (e.g. RGGB to BGGR). Hence, one can normalize any Bayer pattern to an unified one by cropping.

\subsubsection{Testing: Unify via Padding, Disunify via Cropping}

We have shown that one can train a Bayer-pattern-specific network with raw images of different patterns. Moreover, it is possible to denoise images of different patterns with the trained network. Due to the fact that every pixel of the input images needs to be processed, instead of cropping some pixels from the input images, we unify their Bayer patterns via padding some pixels. After network denoising, we simply remove these extra pixels to disunify the output images. This process is illustrated in Fig.~\ref{fig:pad}.

Padding alters the Bayer pattern in a similar way to cropping. Padding one row of pixels to the top and the bottom changes a $C_1C_2C_3C_4$ image into a $C_3C_4C_1C_2$ image (e.g. GRBG to BGGR); padding one column to the left and to the right turns $C_1C_2C_3C_4$ into $C_2C_1C_4C_3$ (e.g. GBRG to BGGR); padding to all four edges transforms $C_1C_2C_3C_4$ into $C_4C_3C_2C_1$ (e.g. RGGB to BGGR).

Hence, we can apply padding to convert any pattern to the desired one. As a straightforward disunification, removing the padded pixels reverses this conversion. Note that we apply reflection padding (aka ``reflect'' for numpy.pad or ``BORDER\_REFLECT\_101'' for OpenCV) to make sure the additional pixels come from the correct channel.

\subsection{Bayer Preserving Augmentation (BayerAug)}
\label{sec:Bayeraug}
\begin{figure}[t]
\begin{center}
\includegraphics[width=\linewidth]{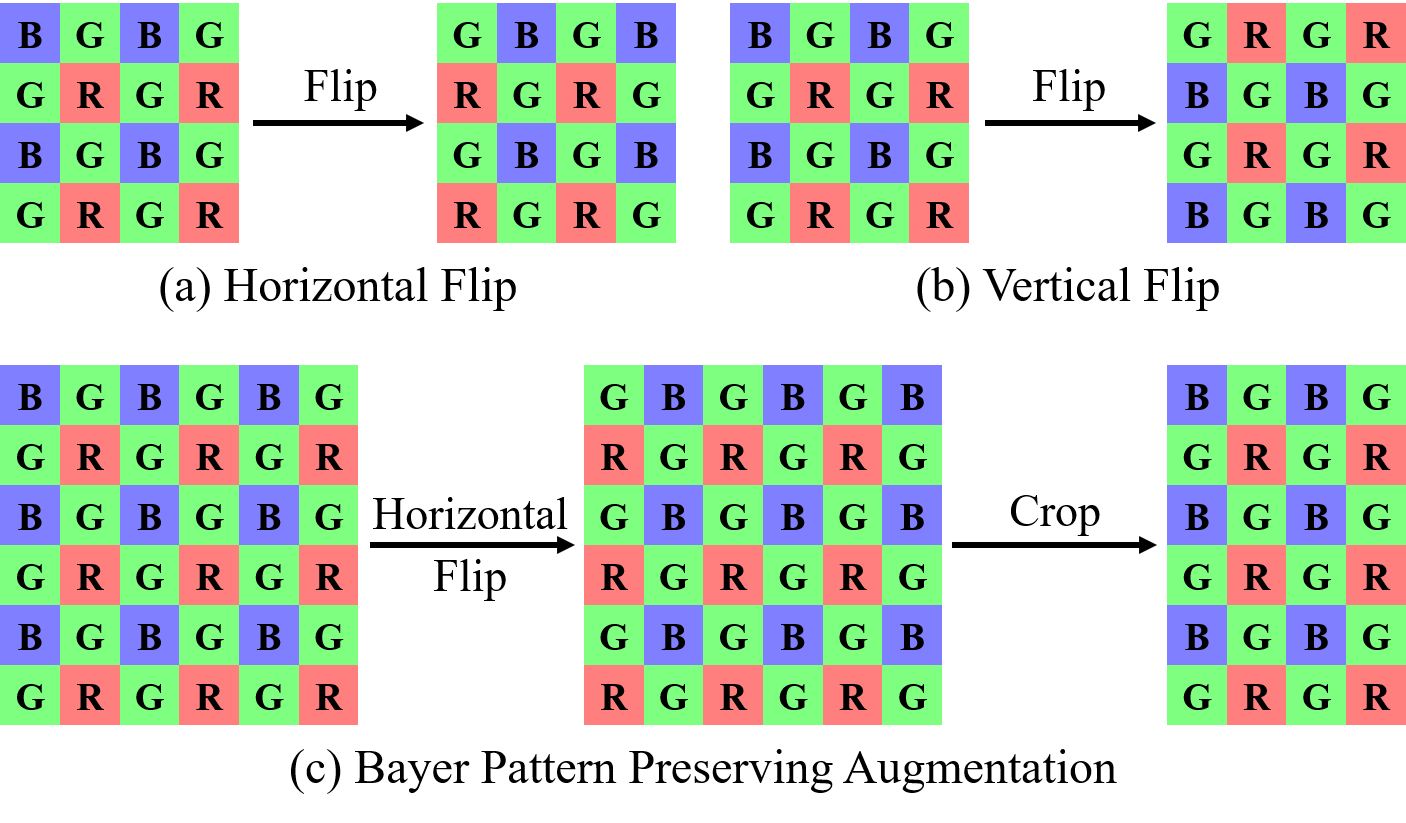}
\end{center}
   \caption{An example of Bayer preserving augmentation. Since flipping an raw image affects its Bayer pattern, to obtain a horizontal flipped BGGR image from a BGGR image, we first perform a horizontal flipping (BGGR$\to$GBRG), and then crop its first and last column (GBRG$\to$BGGR).}
\label{fig:flip_n_crop}
\end{figure}

When training a neural network for vision and graphic tasks on RGB images, it is common to apply flipping and cropping as data augmentation methods. They increases the effective number of samples dramatically while being very concise. However, for Bayer raw images, flipping operations may affect the Bayer pattern. As illustrated in Fig.~\ref{fig:flip_n_crop} (a) and (b), a horizontal flip switches the Bayer pattern from $C_1C_2C_3C_4$ to $C_2C_1C_4C_3$, and a vertical one switches the pattern from $C_1C_2C_3C_4$ to $C_3C_4C_1C_2$.

Therefore, we combine both flipping and cropping to perform data augmentation while preserving the Bayer pattern of the image. After flipping an image, we apply cropping to reverse the change of Bayer pattern. We illustrate this process in Fig. \ref{fig:flip_n_crop} (c).

As another type of flipping, a transposition has different effects on different patterns, depending on the channels of the diagonal components. Generally, the transpose of a $C_1C_2C_3C_4$ image would be in the pattern of $C_1C_3C_2C_4$. For a RGrGbB input, its transpose would be in RGbGrB, which is roughly the same format (assuming the different between Gr channel and Gb channel is subtle). However, for a GRBG input, its transpose would be in GBRG, a totally different pattern. For this reason, we can safely perform transposition to augment RGGB and BGGR images, but not in GRBG or GBRG.

Training with patches instead of the entire images is another common trick used in model training. Different from the cropping operations in BayerUnify, to correctly obtain patches from the entire Bayer raw image without changing its Bayer pattern, we need to avoid any offset. This could be done by simply cropping even numbers of rows (columns). 

With combinations of the discussed three flipping methods and one cropping method, we are able to perform data augmentation on Bayer raw images without any flaw. Note that they can be applied on both homogeneous datasets \cite{chen2018learning} and heterogeneous datasets \cite{SIDD_2018_CVPR}, enhancing the generalizability of the obtained model.

%------------------------------------------------------------------------
\section{Experiment}
\label{sec:experiment}

\subsection{Dataset}
We evaluate our method on the Smartphone Image Denoising Dataset (SIDD) \cite{SIDD_2018_CVPR}. Its training set contains 320 pairs of noise-free images and noisy images, which cover 3 different Bayer patterns and 10 different scenes. Its validation set and testing set consist of 40 pairs of image from 8 different scenes. Both raw images and sRGB images are available.

To compare the generalizability of the obtained models, we need to avoid testing on a trained scene. Therefore, we divided the original SIDD training set into two parts: all ``Scene 1'' image pairs as our testing partition, and the remaining pairs as our training partition. The details are shown in Table~\ref{table:dataset}.

\begin{table}[t]
\begin{center}
\resizebox{\linewidth}{!}{
\begin{tabular}{lccccc}
\toprule
Set  & Scene & \# GRBG & \# BGGR & \# RGGB & Total \\ \midrule
Train & 2-10  & 40      & 126     & 98      & 264   \\
Test  & 1     & 60      & 32      & 20      & 56    \\ \bottomrule
\end{tabular}
}
\end{center}
\caption{Our train/test split of SIDD dataset.}
\label{table:dataset}
\end{table}

\subsection{Network Architecture and Training Details}
\label{sec:network}
Table \ref{table:network} shows the modified U-Net \cite{ronneberger2015u} architecture used in our experiments. As proposed by \cite{chen2018learning}, we packed the raw images into 4 channels as the network input. Differently, we trained the networks to produce 4-channel outputs, and unpacked them to obtain denoised raw images.

In our experiments, all the networks were trained with $L_1$ loss and AdamW optimizer \cite{loshchilov2017fixing} with initial learning rate of $2e-4$ and weight decay of $2e-5$. Patch size and mini-batch size were set to 512 and 4 respectively. We trained each model for $200,000$ iterations, and divided the learning rate by $10$ on plateaus. We detected the plateaus and selected the best models using the PSNR scores on the scene 1 patches of the official validation set. For testing, we fed the entire images to the network, and measured the average PSNR scores of the outputs.

\begin{table}[t]
\begin{center}
\begin{tabular}{lcl}
\toprule
Name  & \# Out & Type \\ \midrule
Input           & 4     & Input Image   \\
EncConv1\_1    & 32    & Conv $3\times3$ \\
EncConv1\_2    & 32    & Conv $3\times3$ \\
Pool1       & 32    & Maxpool $2\times2$ \\
EncConv2\_1    & 64    & Conv $3\times3$ \\
EncConv2\_2    & 64    & Conv $3\times3$ \\
Pool2       & 64    & Maxpool $2\times2$ \\
EncConv3\_1    & 128   & Conv $3\times3$ \\
EncConv3\_2    & 128   & Conv $3\times3$ \\
Pool3       & 128   & Maxpool $2\times2$ \\
EncConv4\_1    & 256   & Conv $3\times3$ \\
EncConv4\_2    & 256   & Conv $3\times3$ \\
Pool4       & 256   & Maxpool $2\times2$ \\
EncConv5\_1    & 512   & Conv $3\times3$ \\
EncConv5\_2    & 512   & Conv $3\times3$ \\
Deconv4     & 256   & Deconv $3\times3$ \\
Add4        & 256   & Deconv4 + EncConv4\_2 \\
DecConv4\_1    & 256   & Conv $3\times3$ \\
DecConv4\_2    & 256   & Conv $3\times3$ \\
Deconv3     & 128   & Deconv $3\times3$ \\
Add3        & 128   & Deconv3 + EncConv3\_2 \\
DecConv3\_1    & 128   & Conv $3\times3$ \\
DecConv3\_2    & 128   & Conv $3\times3$ \\
Deconv2     & 64   & Deconv $3\times3$ \\
Add2        & 64    & Deconv2 + EncConv2\_2 \\
DecConv2\_1    & 64   & Conv $3\times3$ \\
DecConv2\_2    & 64   & Conv $3\times3$ \\
Deconv1     & 32   & Deconv $3\times3$ \\
Add1        & 32    & Deconv1 + EncConv1\_2 \\
DecConv1\_1    & 32   & Conv $3\times3$ \\
DecConv1\_2    & 32   & Conv $3\times3$ \\
DecConv1\_3    & 4   & Conv $3\times3$ \\
Add0        & 4    & Input + DecConv1\_3 \\
\bottomrule
\end{tabular}
\end{center}
\caption{Network architecture used in our experiments. The network takes packed 4-channel raw images as input. Prior to each convolution except EncConv1\_1, a PReLU \cite{he2015delving} is applied as a pre-activation.}
\label{table:network}
\end{table}

\subsection{Results and Analysis}

To show the effectiveness of our proposed BayerUnify and BayerAug, we compared them with naive training methods. The results of the obtained models are shown in Table \ref{table:result1}.

\begin{table}[t]
\begin{center}
%\resizebox{\linewidth}{!}{
\begin{tabular}{lcccc}
\toprule
Method             & GRBG  & BGGR  & RGGB \\ \midrule
GRBG Only          & 43.46 &   -   &   -   \\
BGGR Only          & -     & 49.50 &   -   \\
RGGB Only          & -     &   -   &  51.59  \\
BayerUnify          & 43.92 & 49.88 & 51.85  \\
\textbf{BayerUnify+BayerAug} & \textbf{44.02} & \textbf{49.92} & \textbf{51.95} \\
\bottomrule
\end{tabular}
%}
\end{center}
\caption{PSNR of different methods. As shown, our proposed BayerUnify and BayerAug improve the performance on unseen testing samples.}
\label{table:result1}
\end{table}

\paragraph{GRBG/BGGR/RGGB Only.} As our baseline, we trained one network for each Bayer pattern. Since the number of samples available for training each model is insufficient, this method resulted in a limited performance.

\paragraph{BayerUnify.} Another network was trained with our proposed Bayer pattern unification (Sec.~\ref{sec:Bayernorm}). In the training phase, we applied cropping to convert all 264 training pairs to BGGR format. In the testing phase, we employ padding to unify and cropping to disunify the test cases. Thanks to the increase of training samples, our method outperforms the previous baseline.

\paragraph{BayerUnify+BayerAug.} We further trained a network with both Bayer pattern unification and Bayer preserving augmentation (Sec.~\ref{sec:Bayeraug}). In the training phase, after unification, we augmented the data via flipping and cropping. The result shows that our data augmentation boosted the generalizability of the obtained model, and consequently improved its performance on the unseen scene.

\begin{figure}[t]
\begin{center}
\includegraphics[width=\linewidth,trim={.5cm 0 .5cm 0},clip]{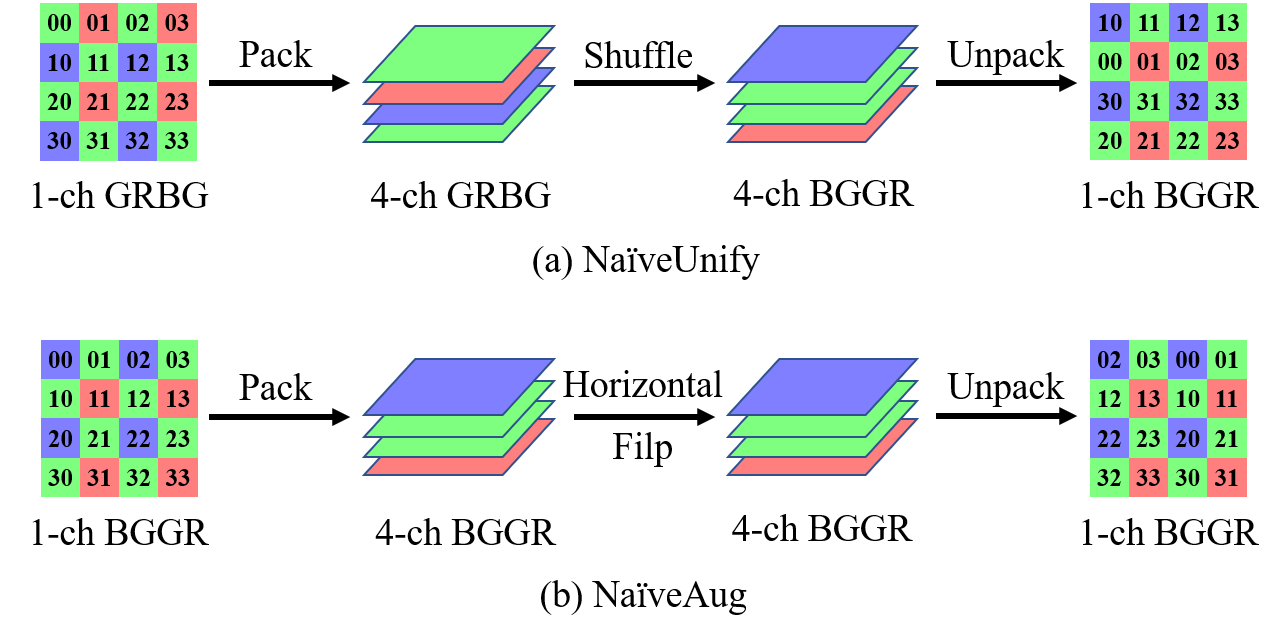}
\end{center}
   \caption{Demonstration of common errors in raw image pre-processing and augmentation. Modifying raw images improperly may disarrange the spatial relationship of the pixels.}
\label{fig:errors}
\end{figure}

Since data pre-processing and augmentation on raw images are not as straightforward as they are on RGB images, it is easy to introduce errors when handling them. Next, we discuss some common errors that we mention in Fig.~\ref{fig:demo}, and show how they dampen the performance. The details of these inappropriate modifications are illustrated in Fig.~\ref{fig:errors}, and the results of them are listed in Table~\ref{table:result2}.

\begin{table}[t]
\begin{center}
%\resizebox{\linewidth}{!}{
\begin{tabular}{lcccc}
\toprule
Method             & GRBG  & BGGR  & RGGB  \\ \midrule
Na\"iveUnify          & 42.78 & 49.74 & 51.83 \\
\textbf{BayerUnify}          & \textbf{43.92} & \textbf{49.88} & \textbf{51.85}  \\
BayerUnify+Na\"iveAug   & 43.83 & 49.76 & 51.81   \\
\textbf{BayerUnify+BayerAug} & \textbf{44.02} & \textbf{49.92} & \textbf{51.95} \\
\bottomrule
\end{tabular}
%}
\end{center}
\caption{PSNR of different pre-processing and augmentation methods. As shown, pre-processing and augmenting raw images via problematic methods (Na\"iveUnify and Na\"iveAug) result in degradation of the network performance.}
\label{table:result2}
\end{table}

\paragraph{Na\"iveUnify.} A common error in raw image pre-processing is to permute the order of the packed 4-channel input. Demonstrated in Fig.~\ref{fig:errors}~(a), while converting the format, it disorganizes the structure information in the original image. To evaluate, we ran a model with training and testing data unified (to BGGR) with this method. Compared to BayerUnify, this method obtains a lower performance, which shows the importance of our valid raw data pre-processing method.

\paragraph{Na\"iveAug.} In raw data augmentation, it is plausible to flip the packed 4-channel images as we do to 3-channel RGB images. However, as shown in Fig.~\ref{fig:errors}~(b), it also disarrays the spatial signal and generates images that are very wandered from the original dataset. We validated this augmentation method based on the correctly unified dataset (BayerUnify). As shown in Table~\ref{table:result2}, this method dampens the results instead of improving them.

\begin{figure*}[t]
\begin{center}
\includegraphics[width=0.98\linewidth]{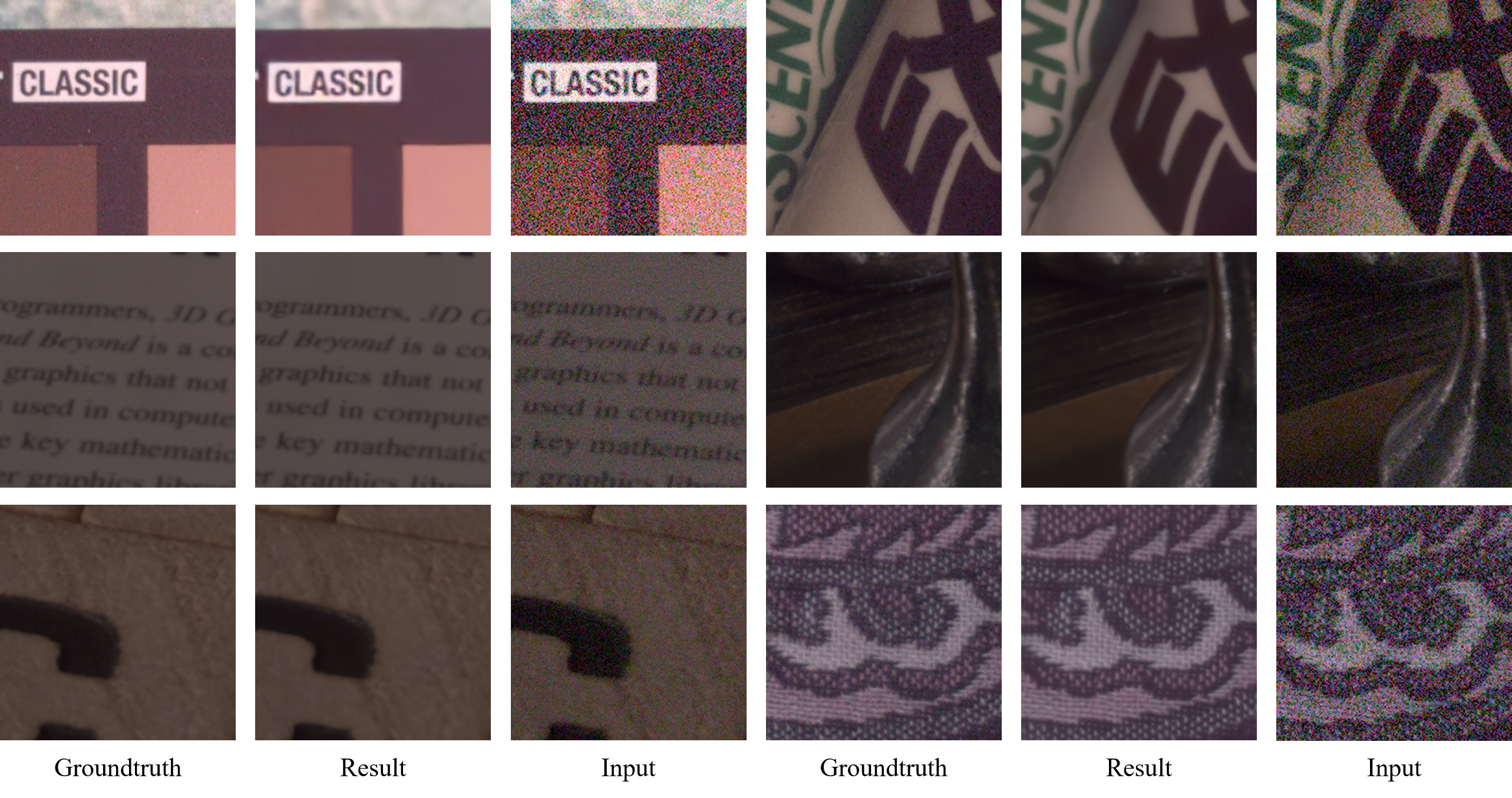} %,trim={.5cm 0 .5cm 0},clip
\end{center}
   \caption{Our results on NTIRE 2019 Real Image Denoising Challenge (Track 1) validation set.}
\label{fig:result}
\end{figure*}

\subsection{Challenge Submission}

After validating our proposed BayerUnify and BayerAug with a specific train-test separation and a relatively small network, we present our solution in NTIRE 2019 Real Image Denoising Challenge (Track 1)~\cite{Abdelhamed_2019_CVPR_Workshops}. We applied our proposed methods together with an enhanced network architecture and a model ensembling strategy. Based on the network we mentioned in Sec.~\ref{sec:network}, we increased the network complexity by enlarging the width of earlier stages, and replacing vanilla convolutional layers with residual blocks ~\cite{he2016identity} (with PReLU, without BN~\cite{ioffe2015batch}). The network architecture is shown in Table~\ref{table:network2}. In the training stage, we split the 320 training pairs into different categories and trained a model for each one, with both BayerUnify and BayerAug. In the testing stage, we performed BayerUnify to normalize the input patches into BGGR images, classified them according to their metadata, and inferred them using the corresponding models. Our proposed solution achieved a PSNR of 52.11 and a SSIM of 0.9969 on the official test set. We show our results in validation set in Fig.~\ref{fig:result}.

\begin{table}[t]
\begin{center}
\begin{tabular}{lcl}
\toprule
Name  & \# Out & Type \\ \midrule
Input           & 4     & Input Image   \\
EncBlock1    & 256    & ResBlock \\
Pool1       & 256    & Maxpool $2\times2$ \\
EncBlock2    & 512    & ResBlock \\
Pool2       & 512    & Maxpool $2\times2$ \\
EncBlock3    & 512   & ResBlock \\
Pool3       & 512   & Maxpool $2\times2$ \\
EncBlock4    & 512   & ResBlock \\
Pool4       & 512   & Maxpool $2\times2$ \\
EncBlock5    & 512   & ResBlock \\
Deconv4     & 512   & Deconv4 $3\times3$ \\
Add4        & 512   & Deconv4 + EncBlock4 \\
DecBlock4    & 512   & ResBlock \\
Deconv3     & 512   & Deconv $3\times3$ \\
Add3        & 512   & Deconv3 + EncBlock3 \\
DecBlock3    & 512   & ResBlock \\
Deconv2     & 512   & Deconv $3\times3$ \\
Add2        & 512    & Deconv2 + EncBlock2 \\
DecBlock2    & 512   & ResBlock \\
Deconv1     & 256   & Deconv $3\times3$ \\
Add1        & 256    & Deconv1 + EncBlock1 \\
DecBlock1    & 256   & ResBlock \\
PReLU    & 256       & PReLU \\
DecConv    & 4   & Conv $3\times3$ \\
Add0        & 4    & Input + DecConv \\
\bottomrule
\end{tabular}
\end{center}
\caption{Network architecture used in our NTIRE 2019 Real Image Denoising Challenge (Track 1) submission. Compared to Table \ref{table:network}, we widened the early stages of the network, and replaced convolutional layers with residual blocks.}
\label{table:network2}
\end{table}

\section{Conclusion}

We have presented effective data pre-processing and augmentation methods specifically designed for Bayer raw images, namely BayerUnify and BayerAug. Our results show that the proposed methods ensembled with advanced network architecture can achieve state-of-the-art performance in raw image denoising problem. We also believe raw image processing with deep learning techniques is a promising direction. 

{\small
\bibliographystyle{ieee_fullname}
\bibliography{egbib}
}

\end{document}